%% file: bmvc_final.tex
\documentclass{bmvc2k}
\usepackage{amssymb}

\title{SceneMiner: Identity-Preserving Multi-Task Fine-Tuning for Unified BEV Scene Mining}

\addauthor{Abdalmalek Aburaddaha}{abdmalek@umich.edu}{1}
\addauthor{Venkatraman Narayanan}{venn@umich.edu}{1}
\addauthor{Keval Thaker}{tkeval@umich.edu}{1}
\addauthor{Samir A. Rawashdeh}{srawa@umich.edu}{1}

\addinstitution{
 University of Michigan-Dearborn\\
 4901 Evergreen Rd\\
 Dearborn, MI 48128, USA
}
\runninghead{Aburaddaha, Narayanan, Thaker, Rawashdeh}{SceneMiner BEV Scene Mining}

\begin{document}

\maketitle

\input{sec/01_abstract}
\input{sec/02_intro}
\input{sec/03_related}
\input{sec/04_method}
\input{sec/05_experiments}
\input{sec/06_results}

\input{sec/07_discussion_conclusion}

\bibliography{egbib}
\end{document}

%% file: sec/01_abstract.tex
\begin{abstract}
Mining hard, safety-critical scenes from driving logs is bottlenecked by
the absence of difficulty labels, and no single proxy, collision risk,
trajectory ambiguity, or semantic rarity suffices to find such scenes
on its own.
We present \textbf{SceneMiner}, a unified, camera-only bird's-eye-view
pipeline that emits complementary mining signals from a frozen
vision-language backbone in a single forward pass, with no LiDAR or
radar: a retrieval embedding for text-prompted scenario search, a
multi-label scene-tag distribution, and a continuous physics-based risk
score (a motion forecast is a byproduct, not a contribution).
Building such a multi-head model exposes our central finding, a failure
mode we term \emph{cross-task interference}: adding or upgrading one head
shifts a shared activation stream and degrades weight-frozen sibling
heads, so freezing parameters alone is insufficient.
Our contribution, \emph{identity-preserving multi-task fine-tuning},
removes this interference by zero-initializing every new sub-module and
freezing every parameter that feeds the shared stream.
The mining heads are thereby preserved bit-identically while training
only ${\sim}102$k parameters.
The tagging head reaches mAP $0.4614$ (micro-F1 $0.5557$) on $20$ scene
tags by pooling each scene into $32$ visual tokens, and the embedding
head supports text-prompted retrieval, validated qualitatively.
Code is available
\href{https://anonymous.4open.science/r/sceneminer_anonymous-64E5}{here}.
\end{abstract}

%% file: sec/02_intro.tex
\section{Introduction}
\label{sec:intro}

Autonomous vehicles must behave reliably across a long-tailed
distribution of driving situations, and the tail, occluded
pedestrian crossings, unprotected left turns, aggressive
cut-ins, contains many of the safety-critical cases that drive crash
risk~\cite{Singh2015NHTSA}.
Establishing statistical confidence in this tail through unguided road
testing is infeasible: demonstrating safety by exposure alone would
require hundreds of millions to billions of
miles~\cite{Kalra2016RAND}.
The practical alternative is to \emph{mine} hard scenes from logs
already collected, to search a corpus for the rare, safety-critical
situations worth re-simulating, labelling, or training on.
This is the problem SceneMiner addresses.
The obstacle is that large-scale corpora such as
nuScenes~\cite{caesar2020nuscenes} ship without ground-truth
difficulty labels, and critically no single proxy can recover them:
forecasting error conflates genuine danger with benign trajectory
ambiguity, language retrieval is blind to multi-agent kinematics, and
physics-based risk heuristics ignore semantic context.
Hard scenes are thus multi-faceted, and a useful miner must emit
\emph{multiple complementary mining signals computed by the same model
on the same input}, so that scores remain directly comparable and a
scene can be flagged hard for any of several reasons at once.

Four mining signals together span the space of difficulty:
(i) a \emph{retrieval embedding} for text-prompted scenario search,
(ii) a \emph{multi-label tag distribution} over weather, road class,
object, and behaviour, (iii) a \emph{scalar risk score} from
proximity, time-to-collision (TTC), and vulnerable-road-user (VRU)
density, and (iv) a byproduct \emph{motion forecast} that exercises the
shared latent and provides a kinematic-difficulty stress test.
SceneMiner produces all four from one frozen
SigLIP2~\cite{tschannen2025siglip2}$+$BEVFormer~\cite{li2022bevformer}
camera-only backbone in a single $204.5$\,ms forward pass on an A40,
with no LiDAR or radar, through a shared $32$-query
Q-Former~\cite{li2023blip2} that pools $32$ visual tokens from BEV
spatial features, camera features, and the motion head's
\texttt{decoded\_tokens} stream.
The entire miner trains only $102{,}031$ parameters, making it far more
parameter-efficient than unified driving
stacks~\cite{hu2023uniad,jiang2023vad} and large VLM-for-driving
systems~\cite{sima2024drivelm}, while supporting dense latent retrieval
where those systems rely on an offline tracker or text-emitting question
answering.

Constructing such a multi-head miner exposes the central
methodological finding of this paper: a failure mode we term
\emph{cross-task interference}, the main obstacle to assembling a
unified miner.
Adding or upgrading one head shifts the activation distribution of a
stream that other, parameter-frozen heads consume through the shared
Q-Former, degrading those heads despite no weight updates.
Concretely, replacing the motion head, a deliberately minimal warm-start
component, with a stronger decoder makes the tag and risk heads degrade
sharply: tag macro-F1 falls by about a third and risk Pearson $r$ flips
sign (Tab.~\ref{tab:ablation}).
This is not classical catastrophic
forgetting~\cite{kirkpatrick2017ewc}, in which weights drift: here the
affected weights are bit-identical, and the harm travels entirely
through the \emph{inputs} of the frozen modules.
In short, frozen weights do not imply frozen activations, so parameter
freezing, the standard multi-task safeguard, is necessary but not
sufficient.

We resolve this interference with \emph{identity-preserving multi-task
fine-tuning}, a discipline that makes head additions safe by
construction.
It couples two requirements: (a) every newly added sub-module is
zero-initialized so the upgraded head is exactly the identity at
warm start, and (b) every parameter that can influence a shared stream
is frozen in the motion head; this leaves only the velocity head and
trajectory projection trainable ($102{,}031$ parameters, under
$10^{-4}$ the roughly $3$-billion-parameter model).
Together these hold the shared activation distribution invariant, so
the addition does not interfere at step zero.
The resulting model (Tab.~\ref{tab:ablation}) keeps the mining
heads---tag micro-F1 ($0.5557$), macro-F1 ($0.2605$), and the risk
head's correlation with its physics target ($r{=}{+}0.3925$)---bit-identical
to the warm-start (within $\pm 10^{-4}$, reproduced across three seeds)
while the byproduct motion head remains functional.
Motion forecasting is not a contribution of this paper: the
$102$k-parameter head is reported only to show that a byproduct head can
be attached to the shared latent without disturbing the tagging, risk,
and retrieval outputs that constitute the mining signals.

\noindent\textbf{Contributions.}
\textbf{(i)}~A unified, camera-only BEV mining pipeline that emits a
retrieval embedding, a $20$-tag distribution, a risk score, and a
byproduct motion forecast in one forward pass using neither LiDAR nor
radar; it attains a tagging mAP of $0.4614$ while training only
$102{,}031$ parameters, two-to-four orders of magnitude fewer than
comparable unified-AD and VLM-for-AD systems.
\textbf{(ii)}~The identification of \emph{cross-task interference}---an
activation-level failure distinct from gradient conflict and weight
drift, and \emph{identity-preserving multi-task fine-tuning}, which
resolves it by zero-initialising new sub-modules and strictly freezing
every parameter that feeds a shared stream so that a head can be added
while the others remain bit-identical to the warm-start
($\Delta\!<\!10^{-4}$ across three seeds).
\textbf{(iii)}~Deterministic, auditable pseudo-labels requiring no human
difficulty annotation: a physics risk score and a tag vocabulary built
from the union of nuPrompt~\cite{wu2025nuprompt} prompts and official
Motional scene descriptions.
\textbf{(iv)}~A non-neural, auditable tag-centroid retrieval pipeline
whose rankings are complementary to those of CLIP on our query sample
($0/15$ top-5 overlap), rather than re-deriving them.

%% file: sec/03_related.tex
\section{Related Work}
\label{sec:related}

\noindent\textbf{Scenario mining and difficulty estimation.}
RefAV~\cite{davidson2025refav} casts scenario mining as a text-to-scene
retrieval problem but does so via an \emph{offline} pipeline: it runs a
3D multi-object tracker and then synthesizes executable filtering code
with an LLM to select matching scenarios.
This pipeline is expressive but costly: it depends on a separate
tracking stage and on per-query code generation, neither of which runs
within a single model pass.
SceneMiner instead performs dense \emph{latent} retrieval directly: a
text query is resolved against $256$-d scene embeddings produced in the
same single forward pass that emits the tagging and risk signals, with
no tracker and no per-query code synthesis.
NuPrompt~\cite{wu2025nuprompt} contributes $35$k+ object-level prompts
on nuScenes for language-conditioned tracking and serves both as a
tag-enrichment source and as our scene-tagging baseline, while
nuScenes-MQA~\cite{inoue2024nuscenesmqa} adds markup-based QA pairs.
These approaches couple mining to a human or LLM annotator; ours does
not. Its retrieval pipeline is instead non-neural and substring-grounded
against the official Motional scene descriptions, giving auditable
provenance. On our query sample, its top-5 results share no scenes with
those of CLIP (zero Jaccard overlap), indicating that the two rankings
are complementary.
Multi-dataset autonomous-driving benchmarks (nuScenes,
Argoverse-2~\cite{wilson2023av2}, Waymo Open~\cite{sun2020waymo})
support cross-domain validation; such transfer is part of our planned
validation rather than a current claim.

\noindent\textbf{BEV perception and unified AD stacks.}
Bird's-eye-view representations have become the dominant substrate for
camera-only perception; BEVFormer~\cite{li2022bevformer} constructs a
unified BEV through spatial-temporal cross-attention and forms the
basis of our backbone.
The broader camera-perception literature continues to advance the
upstream signals such a substrate is built on, spanning multimodal
detection fusion~\cite{narayanan2026mambafusion}, unsupervised scene-flow
estimation~\cite{ahuja2024optflow}, and position encodings for
wide-field-of-view cameras~\cite{ahuja2026fishrope}; SceneMiner is
agnostic to these choices and simply consumes a frozen BEV feature map.
Unified stacks such as UniAD~\cite{hu2023uniad},
VAD~\cite{jiang2023vad}, and BEVerse~\cite{zhang2022beverse} jointly
learn perception, prediction, and planning under a single feature
graph.
We share their multi-task philosophy but differ on \emph{purpose} and
on \emph{footprint}: SceneMiner targets scene-mining
signals—retrieval, tags, and risk, rather than planning—and emits them
from only $102{,}031$ trainable parameters in a single $204.5$\,ms pass.
It is faster than UniAD and latency comparable to VAD-Base, while
training orders of magnitude fewer parameters than UniAD, VAD, and
DriveLM-Agent~\cite{sima2024drivelm} (Tab.~\ref{tab:effsota}).
Dedicated forecasters span graph-structured dynamics
(Trajectron++~\cite{salzmann2020trajectron}), heterogeneous-input
attention (Wayformer~\cite{nayakanti2023wayformer}), goal-anchored
decoding (MTR~\cite{shi2022mtr}), and waypoint-token language modeling
(MotionLM~\cite{seff2023motionlm}).
We make no forecasting claim and do not compete on minADE: SceneMiner's
motion head is a deliberately compact byproduct of the shared Q-Former,
reported only to show that it is functional and non-interfering. This
literature is relevant here only because the pressure to upgrade such a
head is what exposes our interference finding.

\noindent\textbf{Vision-language models for autonomous driving.}
DriveLM~\cite{sima2024drivelm}, DriveVLM~\cite{tian2024drivevlm}, and
nuScenes-MQA~\cite{inoue2024nuscenesmqa} bring VLMs to AD question
answering and captioning, and Qwen2.5-VL~\cite{bai2025qwen25vl} offers a
capable open backbone.
A key distinction for mining is the \emph{output type}: VQA-class
methods emit free-form text, which a downstream system must re-parse and
which cannot be indexed or compared as a vector.
SceneMiner instead emits a retrievable $256$-d embedding alongside
discriminative tag and risk heads, so a corpus can be indexed once and
queried densely.
Talk2BEV~\cite{dewangan2024talk2bev} grounds language in BEV but, like
the QA systems, is oriented toward dialogue rather than vectorized
retrieval.
SceneMiner adopts the BLIP-2 Q-Former~\cite{li2023blip2} as a 32-query
bridge over its BEV, camera, and motion streams; here, the Q-Former is
supervised purely through the task heads, without language modeling
fine-tuning on top of these queries.

\noindent\textbf{Multi-task interference and function-preserving expansion.}
Continuous learning methods such as EWC~\cite{kirkpatrick2017ewc} and
LwF~\cite{li2017lwf} preserve knowledge across sequential tasks by
regularizing weights, while PCGrad~\cite{yu2020pcgrad} and
CAGrad~\cite{liu2021cagrad} resolve \emph{conflicting gradients} in
joint multi-task training.
Gradient-conflict diagnostics, e.g.,\ cosine similarity between task
gradients, reveal when tasks conflict \emph{during} optimization.
The interference we study is fundamentally different in locus: it is
\emph{distributional} and arises \emph{at warm-start}, before any
gradient step, because the inputs to frozen modules shift rather than
their parameters.
The standard mitigation is parameter freezing, often with
adapters~\cite{houlsby2019adapters}, LoRA~\cite{hu2022lora}, or
side-tuning~\cite{zhang2020sidetuning}; identity-preserving expansion
(zero-initializing new layers so the enlarged network is the identity at
warm-start) appears in ControlNet~\cite{zhang2023controlnet}
(diffusion conditioning), adapters and LoRA with $B{=}0$
\cite{houlsby2019adapters,hu2022lora}, and
Net2Net~\cite{chen2016net2net} (layer widening).
All of these operate at the \emph{weight} level and tacitly assume that
frozen modules stay correct under expansion—an assumption our setting
violates. The affected heads' weights are bit-identical across runs, yet
the motion head's \texttt{decoded\_tokens} shift when its architecture
changes, and this activation-distribution shift propagates through a
frozen Q-Former and corrupts the tag and risk heads (Tab.~\ref{tab:ablation}).
Our remedy combines identity initialization with a \emph{strict freeze of
every parameter feeding the shared stream}. To our knowledge, this
pairing has not been isolated and validated in prior multi-head BEV
work, and it addresses an activation-level defect orthogonal to both
gradient-conflict MTL and catastrophic forgetting.

%% file: sec/04_method.tex
\section{Method}\label{sec:method}

\begin{figure*}[t]
\centering
\includegraphics[width=\textwidth]{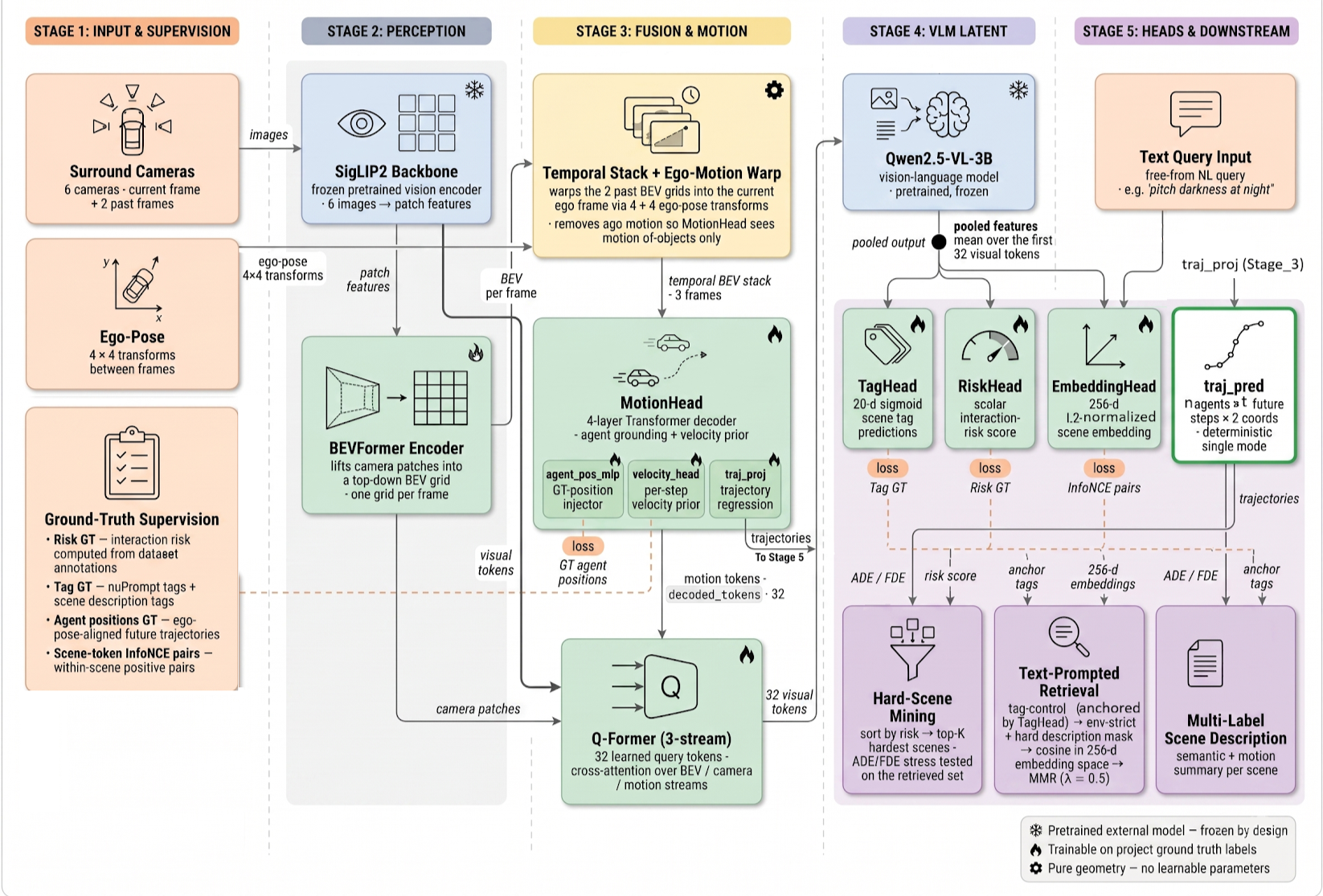}
\caption{SceneMiner architecture. A frozen
SigLIP2~\cite{tschannen2025siglip2} backbone and a
BEVFormer~\cite{li2022bevformer} encoder lift the six surround-view
cameras into a bird's-eye-view grid; a temporal stack and the Motion
Head produce per-agent \texttt{decoded\_tokens}, which a 32-query
Q-Former~\cite{li2023blip2} pools together with the BEV and camera
streams. The pooled tokens feed three parallel heads—tagging, risk,
retrieval embedding, and a byproduct trajectory head, whose outputs
drive the downstream mining tasks. Snowflakes mark frozen pretrained
modules, flames mark the components trained on project labels, and the
gear marks the parameter-free temporal warp.}
\label{fig:arch}
\end{figure*}

\subsection{Overview}\label{sec:method:overview}

SceneMiner is a single camera-only model that ingests six
surround-view RGB streams and emits four scene-mining outputs in a
single forward pass: per-agent future trajectories, a scalar danger
score, a 20-way multi-label tag vector, and a 256-dimensional
retrieval embedding (Fig.~\ref{fig:arch}).
The model is organized as three shared stages feeding three parallel
task heads.
The shared stages are (i)~a frozen SigLIP2~\cite{tschannen2025siglip2}
visual encoder, (ii)~a BEVFormer-style~\cite{li2022bevformer}
spatial--temporal encoder that lifts per-camera patch features into a
$50{\times}50$ bird's-eye-view (BEV) grid, and (iii)~a 32-query
Q-Former~\cite{li2023blip2} that pools the BEV, camera, and motion
streams into a compact set of scene tokens shared by the heads.
The methodological contribution of this paper is
\emph{identity-preserving multi-task fine-tuning}
(Sec.~\ref{sec:method:protocol}), which we apply here to upgrade the
architecture of the motion head while leaving the other three heads
unchanged.
We retain the Q-Former because it is the intended bridge to a
vision language model: the per-agent latents that the Motion Head
emits, denoted \texttt{decoded\_tokens}, flow through the Q-Former to
all three heads, and the Q-Former is the natural attachment point for
caption-level supervision (Sec.~\ref{sec:method:qformer}), which we do
not train.

\subsection{BEV Feature Extraction}\label{sec:method:bev}

\noindent\textbf{Camera encoding and BEV lifting.}
The first stage encodes the cameras and lifts them into a common
bird's-eye-view representation.
The six surround-view cameras are passed through a frozen SigLIP2
backbone~\cite{tschannen2025siglip2}, which yields per-camera patch
embeddings $F^{\mathrm{cam}}_i\!\in\!\mathbb{R}^{N_{p}\times d_{\mathrm{cam}}}$.
A BEVFormer-style spatial cross-attention (SCA) encoder then lifts these
features into a unified $50{\times}50$ BEV grid through the standard
deformable sum~\cite{li2022bevformer}. For a BEV query $\mathbf{Q}_p$
at grid position $p$, the encoder computes
\begin{equation}
\mathrm{SCA}(\mathbf{Q}_p, F^{\mathrm{cam}})
=\frac{1}{|V_{\mathrm{hit}}(p)|}\!\!
\sum_{i\in V_{\mathrm{hit}}(p)}\;\sum_{j=1}^{N_{\mathrm{ref}}}
\mathrm{DeformAttn}\!\bigl(\mathbf{Q}_p,\mathbf{p}_{ij},F^{\mathrm{cam}}_i\bigr),
\label{eq:sca}
\end{equation}
where the index $i$ ranges over the cameras whose viewing frusta cover
$p$ (the set $V_{\mathrm{hit}}(p)$), the index $j$ ranges over the
$N_{\mathrm{ref}}$ deformable sampling points projected into camera
$i$, and $\mathbf{p}_{ij}$ is the reference location of the $j$-th
sampling point in the $i$-th camera. Dividing by $|V_{\mathrm{hit}}(p)|$
averages the contributions of every camera that observes the cell, so a
cell seen by several cameras is not over-counted.
The resulting feature volume
$F^{\mathrm{bev}}\!\in\!\mathbb{R}^{B\times 2500\times 256}$ is a
$50{\times}50$ grid of $256$-channel cells at $2.048$\,m resolution,
covering a $102.4{\times}102.4$\,m ego-centric square.

\noindent\textbf{Temporal stack.}
The second part of this stage adds kinematic context by stacking
several BEV maps over time.
We use a $T{=}3$ stack of BEV maps (the current keyframe and the two
preceding ones, spaced $\Delta t{=}0.5$\,s apart).
Because the ego vehicle moves between keyframes, the two past BEVs are
first SE(2)-warped into the current ego frame using the nuScenes
\texttt{ego\_pose} record. The aligned maps are then concatenated along
the feature dimension and projected back to $D{=}256$ by a
$1{\times}1$ convolution, \texttt{temporal\_proj}:
\begin{equation}
F^{\mathrm{bev}}_{T}
=
\mathrm{temporal\_proj}\!\bigl(
[F^{\mathrm{bev}}_{t},\,
\mathcal{W}_{t-1\to t}(F^{\mathrm{bev}}_{t-1}),\,
\mathcal{W}_{t-2\to t}(F^{\mathrm{bev}}_{t-2})]
\bigr).
\label{eq:temporal}
\end{equation}
Here $\mathcal{W}_{t-k\to t}$ denotes the warp from keyframe $t{-}k$ into
the current frame. The past BEVs are produced under
\texttt{torch.no\_grad}, so only \texttt{temporal\_proj} is trainable,
and it is initialised so that it copies the current-frame slice
unchanged (the identity initialisation of
Sec.~\ref{sec:method:protocol}).

\subsection{Q-Former Tokenisation}\label{sec:method:qformer}

The third shared stage compresses the BEV, camera, and motion streams
into a fixed-size set of scene tokens that the heads can consume
cheaply. The Q-Former~\cite{li2023blip2} pools these three streams into
$32$ scene tokens shared by the Risk, Tagging, and Embedding heads.
A set of $K{=}32$ learnable queries $\mathbf{Q}_{\mathrm{learn}}$
cross-attends to the concatenation of the three streams:
\begin{equation}
Z \;=\; \mathrm{CrossAttn}\!\bigl(
\mathbf{Q}_{\mathrm{learn}},\;
[F^{\mathrm{bev}}_{T};\,F^{\mathrm{cam}};\,\Pi_{m}(h)]
\bigr)
\in\mathbb{R}^{B\times 32\times D},
\label{eq:qformer}
\end{equation}
where $h\!\in\!\mathbb{R}^{B\times N_{a}\times D}$ is the Motion
Head's per-agent decoder output (Sec.~\ref{sec:method:motion}) and
$\Pi_{m}$ is a linear projection.
The third stream, \texttt{decoded\_tokens}, is what couples the Motion
Head to the Q-Former: it is at once an output of the Motion Head and an
input to the Q-Former. Any change to the Motion Head therefore perturbs
the distribution of the pooled tokens $Z$, and with it the features that
the three pooled-token heads consume.
This coupling is the reason the training protocol of
Sec.~\ref{sec:method:protocol} is needed. The central observation is
that freezing the parameters of a downstream head does not freeze its
inputs: modifying an upstream module that feeds a shared stream shifts
the activation distribution that reaches the frozen heads, even though
their weights are untouched, and heads tuned to the previous
distribution degrade. A redesigned Motion Head can thus corrupt
weight-frozen downstream heads through the \texttt{decoded\_tokens}
stream---an effect we quantify in Sec.~\ref{sec:method:protocol}.

\subsection{Four Parallel Heads}\label{sec:method:heads}

Three of the four heads read from the same pooled representation,
whereas the Motion Head reads the BEV directly. Specifically, the Risk,
Tagging, and Embedding heads consume the mean-pooled scene token
$\bar{z}{=}\tfrac{1}{32}\sum_{k}Z_{:,k,:}$, while the Motion Head
consumes the BEV through its own decoder.

\noindent\textbf{Motion Head.}\label{sec:method:motion}
The Motion Head is a transformer decoder that operates over $N_{a}{=}8$
learned agent queries, ranked by detection confidence.
Each query is biased by an MLP applied to the agent's BEV position
$a_a\in\mathbb{R}^{2}$, which anchors the prediction to that agent's
location:
\begin{equation}
q_a \;=\; \mathbf{Q}^{\mathrm{agent}}_a + \mathrm{agent\_pos\_mlp}(a_a),
\label{eq:agentq}
\end{equation}
where $\mathrm{agent\_pos\_mlp}$ supplies explicit BEV grounding.
The 4-layer decoder (each layer applying self-attention, BEV
cross-attention, and a feed-forward block) outputs per-agent latents
$h_a\in\mathbb{R}^{D}$; these latents are the \texttt{decoded\_tokens}
of Eq.~\ref{eq:qformer}.
Two linear sub-heads read $h_a$: \texttt{velocity\_head} predicts a
velocity $v_a\in\mathbb{R}^{2}$ and \texttt{traj\_proj} predicts a
$6$-step residual sequence $\{r_{a,t}\}_{t=0}^{5}$. The two are
recombined into trajectories by
\begin{equation}
y_{a,t} \;=\; a_a + (t+1)\,\Delta t\,v_a + r_{a,t},
\qquad t\in\{0,\dots,5\},
\label{eq:velres}
\end{equation}
with $\Delta t{=}0.5$\,s.
This additive factorisation has a property we exploit in
Sec.~\ref{sec:method:protocol}: when the residual vanishes
($r\!\equiv\!\mathbf{0}$) the head reduces exactly to a
constant-velocity forecaster, and when the velocity vanishes as well it
reduces to the static prediction $y_{a,t}{=}a_a$. These limiting cases
are what make a benign identity initialisation possible.

\noindent\textbf{Risk Head.}
The Risk Head is a two-layer MLP $\mathbb{R}^{D}\!\to\!\mathbb{R}$
applied to $\bar{z}$, with a sigmoid output that produces a scalar
danger score in $[0,1]$. It is supervised by the pseudo-label of
Sec.~\ref{sec:method:pseudo} through a Pearson-style regression loss.

\noindent\textbf{Tagging Head.}
The Tagging Head is a single linear map
$\mathbb{R}^{D}\!\to\!\mathbb{R}^{20}$ over $\bar{z}$, trained with a
multi-label binary cross-entropy loss over the 20-tag nuPrompt
vocabulary~\cite{wu2025nuprompt} (scene, object, and behaviour groups).

\noindent\textbf{Embedding Head.}\label{sec:method:embed}
The Embedding Head is a two-layer MLP that projects the pooled latent
through a $512$-d hidden bottleneck to a $256$-d
$\ell_{2}$-normalised embedding. It is supervised by an InfoNCE
loss~\cite{oord2018cpc} that treats keyframes of the same scene as
positives:
\begin{equation}
\mathcal{L}_{\mathrm{embed}}
=-\frac{1}{|\mathcal{P}|}\!\sum_{(i,j)\in\mathcal{P}}\!
\log\frac{\exp(e(s_{i})^{\top}e(s_{j})/\tau)}
{\sum_{k\neq i}\exp(e(s_{i})^{\top}e(s_{k})/\tau)},
\label{eq:infonce}
\end{equation}
with $\tau{=}0.07$.
At inference, text queries are mapped to nuPrompt tag synonyms; the
centroid of $e(\cdot)$ over tag-matching scenes is the query vector,
and a strict substring mask over the official nuScenes
\texttt{scene.description} filters environmental candidates.

\subsection{Pseudo-Label Construction}\label{sec:method:pseudo}

Because nuScenes~\cite{caesar2020nuscenes} provides no ground-truth
difficulty labels, we derive two deterministic pseudo-labels from the
annotations alone, using no human labels, no learned uncertainty, and no
model ensemble.

\noindent\textbf{Risk pseudo-label.}
For each keyframe we aggregate three components and then take the
maximum over the keyframes of a scene. The three components capture
distinct precursors of difficulty.
\emph{Proximity} is $\rho_{\mathrm{prox}}\!=\!\exp(-d_{\min}/5\,\mathrm{m})$,
where $d_{\min}$ is the distance from the ego vehicle to the nearest
annotated agent.
\emph{Time-to-collision} (TTC) is the smallest positive
constant-velocity time-to-collision, gated to a closest-approach
distance below $3$\,m and mapped to
$\rho_{\mathrm{ttc}}\!=\!\max(0,1{-}\mathrm{TTC}/4\,\mathrm{s})$.
\emph{Vulnerable-road-user} (VRU) density is the share of pedestrians,
cyclists, and motorcyclists within $30$\,m, normalised by 10.
The pseudo-label combines the three linearly:
\begin{equation}
\hat{r}(s) = 0.4\,\rho_{\mathrm{prox}} + 0.3\,\rho_{\mathrm{ttc}}
+ 0.3\,\rho_{\mathrm{vru}},
\label{eq:risk-label}
\end{equation}
each component clipped to $[0,1]$ ($\hat{r}$ has mean $0.41$, std
$0.18$ across 850 scenes).
The component weights ($0.4$, $0.3$, $0.3$) were tuned on the training
split to maximise the rank correlation between $\hat{r}$ and per-scene
motion difficulty, with proximity weighted most heavily as the most
direct collision precursor.

\noindent\textbf{Tag enrichment.}
The tag pseudo-label is a 20-bit per-scene vector formed by OR-merging
two sources: (a) the per-frame nuPrompt~\cite{wu2025nuprompt} object
prompts, mapped through a synonym list, and (b) the per-scene Motional
\texttt{scene.description} string, tokenised against the same list. The
second source recovers weather, lighting, and infrastructure cues that
nuPrompt alone misses.
After enrichment, $7.3$ tags fire per scene on average and only $2$ of
the $850$ scenes have an empty vector.

\subsection{Identity-Preserving Fine-Tuning}\label{sec:method:protocol}

This subsection presents the methodological core of the paper: how to
upgrade the motion head without disturbing the other heads. We begin
from a multi-head warm start in which the backbone and the four heads
are trained jointly. We then wish to upgrade the motion head, adding a
$T{=}3$ temporal stack (Eq.~\ref{eq:temporal}), the
\texttt{agent\_pos\_mlp} grounding (Eq.~\ref{eq:agentq}), three further
decoder layers, and a velocity subhead, while leaving the other heads
intact.

A direct fine-tune fails at this, even when every other head is held
frozen. The newly initialised layers shift the per-agent latents $h$,
this shift propagates through the shared Q-Former stream
(Eq.~\ref{eq:qformer}), and the heads tuned to the warm-start
distribution degrade sharply (Sec.~\ref{sec:ablation}). The diagnosis is
that the defect lies in the activations, not in the weights, so the
remedy must act on the activations rather than on the weights. We
therefore combine two requirements that together hold
\texttt{decoded\_tokens} bit-identical to the warm-start at step~$0$ and
keep it from drifting thereafter.

\noindent\textbf{(i) Identity initialization.}
The first rule makes the upgraded motion head start out exactly equal to
the warm-start one. Every newly added sub-module is zero-initialized so
that the expanded head reproduces the warm-start output. Concretely, the
output projection of each new decoder layer is zeroed, which makes that
layer the identity under its residual connection, and the final linear
layers of \texttt{agent\_pos\_mlp}, \texttt{temporal\_proj}, and
\texttt{velocity\_head} are zeroed, which reduces Eq.~\ref{eq:velres} to
$y_{a,t}=a_{a}+r_{a,t}$. Under this scheme \texttt{decoded\_tokens}
matches the warm start to within numerical noise.

\noindent\textbf{(ii) Strict freeze of the \texttt{decoded\_tokens}
path.}
The second rule keeps the head equal to the warm start as training
proceeds, by freezing everything that could move the latents $h$. We
freeze the full decoder, the agent queries, \texttt{agent\_pos\_mlp},
\texttt{temporal\_proj}, the BEV encoder, the Q-Former, and the three
sibling heads, leaving only \texttt{velocity\_head} and
\texttt{traj\_proj} trainable ($102{,}031$ parameters, under $10^{-4}$
of the roughly $3$-billion-parameter model). Neither trainable layer
lies on the computation graph that produces \texttt{decoded\_tokens}, so
the shared tokens $Z$ are invariant at initialization by construction.
Empirically they remain so throughout fine-tuning: the mining-head
metrics show zero variance across seeds (supplementary material), which
also reports the full three-stage recipe and hyperparameters.

%% file: sec/05_experiments.tex
\section{Experiments}
\label{sec:experiments}

\noindent\textbf{Dataset and evaluation scope.}
All experiments use a single dataset under a camera-only protocol.
We train and evaluate on nuScenes
v1.0-trainval~\cite{caesar2020nuscenes} under the standard
$700$/$150$ train/val scene split, using only the six surround-view
cameras and no LiDAR or radar at any stage.
Of the $6{,}019$ validation keyframes, the $5{,}820$ that carry at
least one valid future-waypoint annotation form the \emph{well-posed}
set on which all metrics are computed; the remaining frames are
degenerate boundary cases with no annotated agents and are excluded.
Tag supervision uses the $20$-tag nuPrompt
vocabulary~\cite{wu2025nuprompt}, enriched with the official
\texttt{scene.description} strings (Sec.~\ref{sec:method:pseudo}).
Evaluation is confined to this single dataset; cross-corpus transfer
(for example to Argoverse-2 or Waymo) is left to future work
(Sec.~\ref{sec:conclusion}).

\noindent\textbf{Implementation.}
All experiments run on $4{\times}$NVIDIA A40 GPUs ($46$\,GiB each)
under CUDA~12.1 and PyTorch~2.3 with bfloat16 mixed precision, optimised
with AdamW (weight decay $0.01$).
The relevant training stages are Stages~2 and~3.
Stage~2 trains the multi-head baseline with a cosine schedule at
learning rate $5{\times}10^{-5}$ and batch size $2$ per GPU (effective
batch $8$) for $12$ epochs.
Stage~3, the identity-preserving motion fine-tune, trains \emph{only}
the $102{,}031$ parameters of \texttt{velocity\_head} and
\texttt{traj\_proj}, at learning rate $1{\times}10^{-4}$, for a single
epoch with early stopping on validation ADE. Every other parameter that
can influence \texttt{decoded\_tokens} is frozen, so the shared stream
remains bit-identical to the Stage-2 baseline at every optimisation step
(Sec.~\ref{sec:method:protocol}).
The BEV grid is $50{\times}50$ at $2.048$\,m per cell (a
$102.4{\times}102.4$\,m ego-centric square), with a $T{=}3$
ego-motion-compensated temporal stack.

\noindent\textbf{Evaluation protocol.}
\label{sec:experiments:protocol}
Each of the four outputs is scored with the metric standard for its
task.
\emph{Tagging} is scored at a binary-cross-entropy (BCE) decision
threshold of $0.5$ by the micro-averaged F1 (pooling all tag decisions
before computing F1), the macro-averaged F1 (the unweighted mean of
per-tag F1), and the mean average precision (mAP, the mean over tags of
the area under each precision-recall curve).
\emph{Risk} is scored by the Pearson and Spearman rank correlation
between predicted risk and per-keyframe motion error and by the mean
absolute error (MAE) against the physics pseudo-label.
\emph{Motion} is scored by the Average and Final Displacement Error
(ADE, the mean $\ell_2$ distance between predicted and ground-truth
waypoints; FDE, the same distance at the final waypoint), computed
mask-aware over the eight ranked agents and six $0.5$\,s waypoints so
that only valid (agent, timestep) pairs contribute.
\emph{Retrieval} is scored by Recall@$K$ (the fraction of queries whose
top-$K$ results contain a correct scene) against \texttt{scene\_token}
identity, together with a description-grounded environmental-tier
check.
For tagging we additionally report a stricter per-tag calibrated
protocol, whose ground truth comes from substring matches against the
official \texttt{scene.description}; it complements the BCE scores.

%% file: sec/06_results.tex
\section{Results}
\label{sec:results}

\subsection{Cross-Task Interference and Its Resolution}
\label{sec:ablation}

Upgrading the Motion Head corrupts the sibling heads through their
shared activations, and the identity-preserving protocol prevents this.
We isolate each ingredient with a $2{\times}2$ factorial over identity
initialization and the strict freeze (Tab.~\ref{tab:ablation}),
spanning the multi-head \emph{Baseline}, an \emph{Unconstrained}
fine-tune (neither ingredient), a \emph{Frozen-Head} variant (freeze
without identity initialization), an \emph{Id-init only} variant
(identity initialization without the freeze), and the full
\emph{identity-preserving} model (both).

The Unconstrained fine-tune reduces motion ADE by almost an order of
magnitude, but the tag and risk heads degrade sharply and the risk
correlation reverses sign. This is the decisive test of the
interference claim, since the tag and risk weights are frozen
throughout and only the motion-head architecture changes: the harm can
travel only through the activation distribution of
\texttt{decoded\_tokens}, the third stream into the shared Q-Former.
Freezing the tag and risk heads as well, as in the frozen-head variant,
does not help but instead worsens the degradation, confirming that
freezing weights cannot stop interference carried through inputs.
Identity initialisation, by contrast, attacks the activation shift at
its source: the Id-init-only variant lifts the mining heads well above
the Unconstrained and Frozen-Head variants (tag macro-F1 $0.1976$, risk
$r{=}{+}0.2175$), yet it does not restore them because the trainable
upstream path drifts away from the identity during fine-tuning. Only
the full identity-preserving model, which adds the strict freeze that
holds that path in place, recovers the mining heads to within $10^{-4}$
of the baseline while retaining the improved motion. Retrieval R@10 is
identical across all configurations because the Embedding Head's
parameters and inputs are never touched by the motion head edits.

\begin{table*}[t]
\begin{center}
\resizebox{\textwidth}{!}{%
\begin{tabular}{l|cc|rr|rr|r}
\hline
Configuration & Id-init & Lock & ADE $\downarrow$ & FDE $\downarrow$ & Tag micro-F1 $\uparrow$ & Tag macro-F1 $\uparrow$ & Risk $r\,\uparrow$ \\
\hline
Baseline                     & --   & --   & 17.89 & 18.11 & 0.5556 & 0.2604 & $+0.3926$ \\
Unconstrained                & $\times$ & $\times$ &  2.06 &  4.29 & 0.4613 & 0.1635 & $-0.0339$ \\
Frozen-Head                  & $\times$ & part. &  2.17 &  4.35 & 0.2669 & 0.1465 & $-0.1945$ \\
Id-init only                 & $\checkmark$ & $\times$ & 2.08 & 4.28 & 0.4825 & 0.1976 & $+0.2175$ \\
\textbf{Identity-Preserving} & $\checkmark$ & $\checkmark$ & \textbf{2.03} & \textbf{4.20} & \textbf{0.5557} & \textbf{0.2605} & $\mathbf{+0.3925}$ \\
\hline
\end{tabular}%
}
\end{center}
\caption{Effect of the identity-preserving protocol on nuScenes val,
arranged as a $2{\times}2$ factorial over identity initialization and
the strict freeze. ``Id-init'' = identity initialization of new
motion-head components; ``Lock'' = freeze of every parameter
influencing \texttt{decoded\_tokens}. Tag F1 uses the BCE@$0.5$
protocol; risk $r$ is the correlation with the physics pseudo-label.
Retrieval R@10 is $0.0653$ for every configuration and is omitted.
Identity initialization alone (Id-init only) partly restores the mining
heads but does not preserve them; only the full Identity-Preserving
model, which adds the strict freeze, keeps tag and risk at their
preserved level while the byproduct motion head stays functional.}
\label{tab:ablation}
\end{table*}

\subsection{Per-Class Motion}
\label{sec:percls}

The byproduct motion head's error grows as the constant-velocity prior
predicts, diverging by agent class at longer horizons
(Tab.~\ref{tab:percls}). The near-zero error at the first step reflects
the accuracy of the anchor position fed through \texttt{agent\_pos\_mlp}.
Beyond it, pedestrian error grows roughly linearly while vehicle error
grows faster than linearly because the prior's unmodelled curvature
accumulates fastest for fast-moving agents; vehicles, therefore, dominate
the aggregate error.

\begin{table}[t]
\begin{center}
\begin{tabular}{l|r|rrrrrr}
\hline
Class & $n$ agents & $t_1$ & $t_2$ & $t_3$ & $t_4$ & $t_5$ & $t_6$ \\
\hline
Pedestrian & 10{,}749 & 0.07 & 0.46 & 0.90 & 1.33 & 1.77 & 2.21 \\
Vehicle    & 31{,}348 & 0.09 & 1.04 & 2.02 & 3.00 & 3.95 & 4.87 \\
Other      & 25       & 0.10 & 0.17 & 0.23 & 0.32 & 0.34 & 0.40 \\
\hline
\textbf{All} & 42{,}122 & 0.08 & 0.89 & 1.74 & 2.57 & 3.40 & \textbf{4.20} \\
\hline
\end{tabular}
\end{center}
\caption{Per-class mask-aware ADE (m) at the six prediction horizons
$t_1{=}0.5$\,s through $t_6{=}3.0$\,s (steps of $0.5$\,s); $n$ is the
number of valid agents. Vehicles dominate the population and the
long-horizon error budget, while pedestrian errors grow roughly
linearly with the velocity prior.}
\label{tab:percls}
\end{table}

\subsection{Multi-Label Scene Tagging}
\label{sec:pertag}

The tagging head reaches mAP $0.4614$ (micro-F1 $0.5557$, macro-F1
$0.2605$) on the $20$ nuPrompt tags, despite pooling the entire scene
into only $32$ visual tokens from a frozen backbone—evidence that a
compact learned bottleneck retains enough structure for reliable scene
attribution.
The per-tag breakdown (Tab.~\ref{tab:tags}) shows where this accuracy
concentrates: the head is strong on the high-prevalence scene and
object tags that matter most for mining, while the low-prevalence long
tail pulls the macro mean down.
The two macro-F1 figures measure different protocols rather than the
same quantity twice: the value in Tab.~\ref{tab:ablation} uses the
stricter BCE@$0.5$ protocol matched to training, whereas
Tab.~\ref{tab:tags} uses per-tag calibration, which is a better
predictor of downstream retrieval.

\begin{table}[t]
\begin{center}
\begin{tabular}{l|r|r}
\hline
Tag & F1 & Prevalence \\
\hline
\texttt{scene\_rainy}            & 0.920 & 0.177 \\
\texttt{scene\_school\_zone}     & 0.912 & 0.839 \\
\texttt{object\_bicycle}         & 0.898 & 0.823 \\
\texttt{object\_bus}             & 0.894 & 0.780 \\
\texttt{behavior\_static}        & 0.816 & 0.673 \\
\texttt{scene\_dark}             & 0.542 & 0.242 \\
\texttt{scene\_residential}      & 0.481 & 0.409 \\
\texttt{scene\_parking\_lot}     & 0.478 & 0.496 \\
\texttt{scene\_intersection}     & 0.348 & 0.348 \\
\texttt{object\_truck}           & 0.327 & 0.391 \\
\hline
\textbf{Macro-F1 (20 tags)}      & \textbf{0.357} & --- \\
\hline
\end{tabular}
\end{center}
\caption{Per-tag F1 on nuScenes val under the per-tag-calibrated
description-substring protocol (top-10 tags shown; full 20-tag macro
mean below). This protocol differs from the BCE@$0.5$ protocol of
Tab.~\ref{tab:ablation}, which is why its macro-F1 ($0.357$) exceeds
the $0.2605$ reported there.}
\label{tab:tags}
\end{table}

\subsection{Risk as a Difficulty Estimator}
\label{sec:risk}

Predicted risk is useful for mining the difficult tail, but only as a
ranking signal rather than a calibrated danger measure. The head fits
its physics pseudo-label well (MAE $0.1350$, Pearson $r{=}{+}0.3925$
against the label). What matters for mining, however, is whether
predicted risk tracks actual scene difficulty, which we measure as its
correlation with \emph{motion error} a target distinct from the
supervising label. Here the correlation is positive but weak (Pearson
$r = {+}0.233$, $r^{2} = 0.054$), so predicted risk explains only a
small part of per-sample difficulty. At the tail it is nonetheless
informative: the top-$50$ scenes by predicted risk are $2.9{\times}$
harder than the validation baseline, so the head reliably identifies
the difficult cases a mining pipeline needs to retrieve. We frame this as a protocol we establish rather than
a comparison we win: prior risk benchmarks use a different risk
definition, dataset, or output type (discrete severity in
CARScenes~\cite{he2025carscenes}, CARLA risk
anticipation in RiskBench~\cite{kung2024riskbench}, closed-loop scores
in nuPlan~\cite{caesar2021nuplan}), so none is directly comparable to a
continuous physics-based scalar on nuScenes keyframes. The validation
is also not independent of its supervision: risk and the motion head
use the same annotations, so the correlation may partly reflect shared
inputs rather than a property of predicted risk that generalizes beyond
its physics heuristic.

\subsection{Text-Prompted Retrieval}
\label{sec:retrieval}

SceneMiner's contribution to scenario mining is a capability rather
than a single comparable score (Tab.~\ref{tab:retsota}): it retrieves
by matching a query against scene embeddings emitted in the same
forward pass as the other signals, reaching $6.5{\times}$ chance at
R@10 under the strict \texttt{scene\_token} pseudo-label. Prior
systems obtain their results by means that do not produce a reusable
index: RefAV~\cite{davidson2025refav} runs an offline 3D tracker and
uses an LLM to synthesise per-query filtering code, while VQA-style
systems (DriveLM~\cite{sima2024drivelm},
OmniDrive~\cite{wang2025omnidrive},
NuScenes-QA~\cite{qian2024nuscenesqa}) emit free-form text rather than
a retrievable embedding. Their metrics (HOTA-Temporal, VQA accuracy,
CIDEr) cannot be placed on the same scale as dense retrieval, so the
table records a difference in capability, not a common score.

\begin{table}[t]
\begin{center}
\begin{tabular}{l|l|l|r}
\hline
Method & Metric & Output & Value \\
\hline
\textbf{SceneMiner (ours)} & R@1/5/10 & dense latent & \textbf{.0066/.0327/.0653} \\
\textbf{SceneMiner (ours)} & desc-mask & dense latent & \textbf{6/6 PASS} \\
RefAV RefProg$^\dagger$~\cite{davidson2025refav} & HOTA-T & tracker+LLM code & 50.1 \\
RefAV RefClassify$^\dagger$~\cite{davidson2025refav} & HOTA-T & 2D-crop classifier & 17.2 \\
RefAV RefBlind$^\dagger$~\cite{davidson2025refav} & HOTA-T & class-filter & 19.2 \\
NuScenes-QA$^\dagger$~\cite{qian2024nuscenesqa} & VQA acc & text & 60.4 \\
DriveLM-Agent$^\dagger$~\cite{sima2024drivelm} & CIDEr & text & 0.044 \\
OmniDrive$^\dagger$~\cite{wang2025omnidrive} & CIDEr & text & 0.686--0.732 \\
\hline
\end{tabular}
\end{center}
\caption{Text-prompted retrieval and scenario mining. The contribution
is architectural: SceneMiner retrieves via a dense latent embedding in
one forward pass, whereas RefAV needs an offline tracker plus LLM code
synthesis and VQA-class methods emit text, not a retrievable vector.
$\dagger$: protocol/metric not commensurable with dense retrieval
(shown for capability context, not as a common score).}
\label{tab:retsota}
\end{table}

The retrieval pipeline maps a text query through the Embedding Head
to a tag centroid (env-priority synonym anchor, env-strict anchor
pool), applies a strict description-substring mask over the official
nuScenes \texttt{scene.description}, and re-ranks with MMR
($\lambda{=}0.5$).
A set of six one-token and full-sentence queries for ``night'' and
``rain'' returns top-$3$ scenes that, on manual inspection of the
official descriptions, all match the queried attribute, with high
within-attribute cosine similarity. Running off-the-shelf
CLIP~\cite{radford2021clip} over the \emph{same} embeddings yields
rankings that share no scenes with ours ($0/15$ top-5 overlap), so the
learned space is complementary to CLIP rather than a copy of it; this
complementarity is why queries must be anchored inside the learned
space through the nuPrompt vocabulary. This check is qualitative rather
than a quantitative benchmark.

\begin{figure*}[t]
\centering
\includegraphics[width=0.92\textwidth]{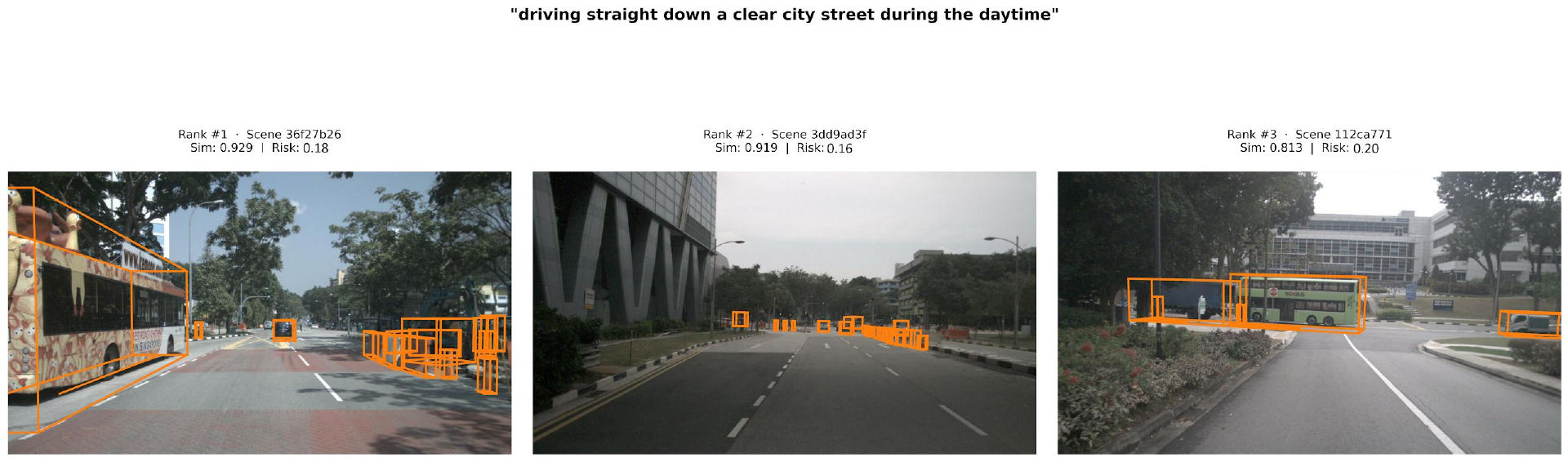}\\[2pt]
\includegraphics[width=0.92\textwidth]{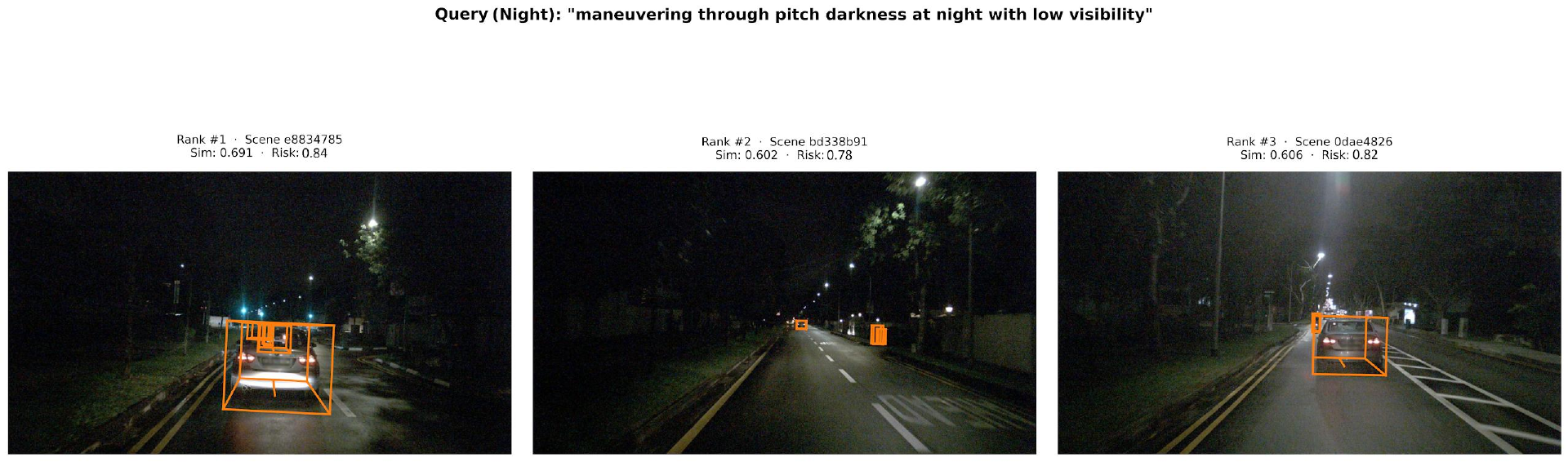}\\[2pt]
\includegraphics[width=0.92\textwidth]{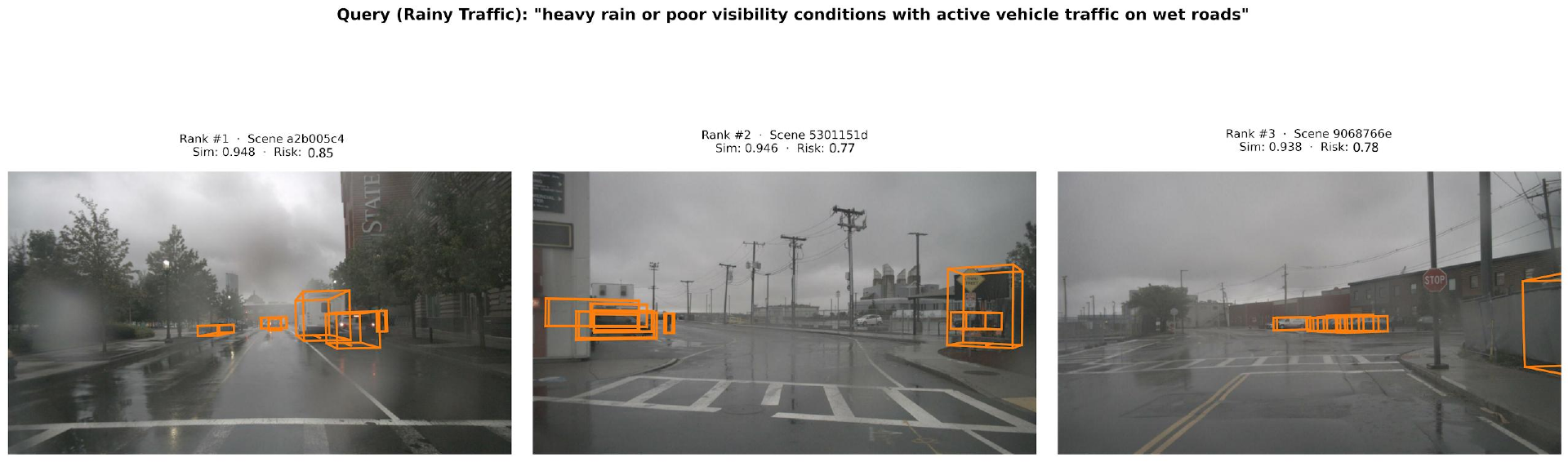}
\caption{Qualitative text-prompted retrieval. Each row shows the
top-$3$ scenes for one query (top: clear daytime traffic; middle:
night with low visibility; bottom: heavy rain on wet roads), with the
cosine similarity and predicted risk of each scene and the detected
agents projected into the front camera. Retrieval remains correct as
conditions worsen, and the predicted risk rises with the difficulty of
the queried condition ($\approx\!0.2$ daytime, $\approx\!0.8$
night/rain), consistent with the risk--difficulty relationship
reported in Sec.~\ref{sec:risk}.}
\label{fig:quali}
\end{figure*}

Fig.~\ref{fig:quali} shows top-ranked retrievals for natural-language
queries spanning easy to hard conditions, with detected agents
projected into the front camera. Two patterns are visible. First,
retrieval stays semantically faithful as conditions degrade: the
daytime, night, and rain queries each return scenes that match the
requested attribute, even though night and rain are the
lowest-prevalence conditions in nuScenes. Second, the predicted risk
attached to each scene tracks query difficulty, near $0.2$ for clear
daytime traffic, around $0.5$ when following a large vehicle at close
range, and above $0.75$ for night and wet-road scenes so the risk
head and the retrieval head agree on which retrieved scenes are
hazardous without being trained to.

\noindent\textbf{Hard-scene mining.}
Running the motion head on the sub-populations each signal retrieves
(Tab.~\ref{tab:hard}) shows that the signals are complementary:
predicted risk picks out the hardest cases overall by dynamic hazard,
while the semantic-retrieval signals pick out different difficult
buckets organised by scene class. A practical mining pipeline can
therefore combine them rather than choose between them.

\begin{table}[t]
\begin{center}
\begin{tabular}{l|r|r}
\hline
Source & ADE (m) & FDE (m) \\
\hline
val baseline ($n {=} 5{,}820$)         & 2.14 & --- \\
top-50 by predicted risk               & \textbf{6.22} & \textbf{12.23} \\
\hline
CLIP query 1 (night)                   & 1.26 & 2.23 \\
CLIP query 2 (intersection)            & 2.26 & 4.48 \\
CLIP query 3 (rain)                    & 4.52 & 9.52 \\
\hline
Ours query 1 (night)                   & 1.81 & 3.36 \\
Ours query 2 (intersection)            & 2.44 & 4.70 \\
Ours query 3 (rain)                    & 1.21 & 2.30 \\
\hline
\end{tabular}
\end{center}
\caption{Motion-head stress under retrieved hard scenes. ADE/FDE here
are averaged \emph{per scene} over the $5{,}820$ well-posed keyframes,
giving a baseline of $2.14$\,m; this differs from the agent-waypoint
mask-aware ADE of $2.03$\,m in Tab.~\ref{tab:ablation}, which averages
over all valid (agent, timestep) pairs. The top-$50$ scenes by
predicted risk reach $2.9\times$ the per-scene baseline, consistent
with risk as a hard-scene-mining signal.}
\label{tab:hard}
\end{table}

\subsection{Inference Efficiency}
\label{sec:hw}

A single forward pass emits all four head outputs in $204.5$\,ms at
$8.6$\,GiB peak memory on one A40 (measured over $20$ warm-up and
$100$ timed passes), with throughput scaling sub-linearly up to batch
$4$. The pipeline is compute-bound on the shared frozen backbone, and
no caption-time vision--language model enters the inference graph.
Against published unified-AD and VLM-for-AD systems
(Tab.~\ref{tab:effsota}), this makes SceneMiner faster than
UniAD~\cite{hu2023uniad} and about on par with
VAD-Base~\cite{jiang2023vad} on latency. The decisive margin, however,
is in trainable parameters: at $102$k parameters SceneMiner trains
two-to-four orders of magnitude fewer than UniAD, VAD, and
DriveLM-Agent~\cite{sima2024drivelm}, even with a full 3B-parameter
backbone in the loop. Baseline figures are quoted from the respective
publications on the hardware each lists; as measurement conditions
differ, the latency comparison is indicative, whereas the parameter
comparison is hardware-independent.

\begin{table}[t]
\begin{center}
\begin{tabular}{l|l|r|r|r}
\hline
Method & HW & Lat.\ (ms)\,$\downarrow$ & FPS\,$\uparrow$ & Train.\ params\,$\downarrow$ \\
\hline
VAD-Tiny~\cite{jiang2023vad}        & A100/3090 & 59.5  & 16.8 & ${\sim}$119\,M \\
MomAD~\cite{song2025momad}          & RTX 4090  & ${\sim}$128 & ${\sim}$7.8 & $>$100\,M \\
\textbf{SceneMiner (ours)} & A40 & \textbf{204.5} & \textbf{4.87} & \textbf{102\,k} \\
VAD-Base~\cite{jiang2023vad}        & A100/3090 & 224.3 & 4.5  & $>$100\,M \\
UniAD~\cite{hu2023uniad}            & A100      & 555.6 & 1.8  & $>$100\,M \\
DriveLM-Agent~\cite{sima2024drivelm} & unspec.  & ${\sim}$2777 & 0.36 & 3.96\,B \\
\hline
\end{tabular}
\end{center}
\caption{Inference efficiency versus published unified-AD and
VLM-for-AD systems. SceneMiner is $2.7{\times}$ faster than UniAD,
latency-comparable to VAD-Base, and trains $980{\times}$--$38{,}800{\times}$
fewer parameters than UniAD/VAD/DriveLM-Agent while running a
3B VLM. Hardware differs per row (each system on its published
hardware); the parameter comparison is hardware-independent.}
\label{tab:effsota}
\end{table}

%% file: sec/07_discussion_conclusion.tex
\section{Conclusion}
\label{sec:conclusion}

We presented SceneMiner, a unified camera-only pipeline for hard-scene
mining that emits a retrieval embedding, scene tags, and a risk score
in a single forward pass, training only ${\sim}102$k parameters.
Its central finding is \emph{cross-task interference}: adding a head
shifts a shared activation stream and degrades weight-frozen sibling
heads. Our \emph{identity-preserving multi-task fine-tuning} removes
this by zero-initializing and freezing the parameters that feed the
stream, a principle that should apply to any multi-head model sharing
a pooled representation.

\noindent\textbf{Future work.}
Three directions follow. (i)~\emph{Cross-dataset transfer}: all
experiments are on nuScenes, so transferring the frozen backbone to
Argoverse-2~\cite{wilson2023av2} and Waymo~\cite{sun2020waymo} would
test the protocol under domain shift. (ii)~\emph{Independent risk
validation}: predicted risk and the motion head share annotations, so
an external ground truth (human-labeled difficulty or an
ensemble-uncertainty baseline) is needed to show the risk signal
generalizes beyond its physics heuristic. (iii)~\emph{Quantitative
retrieval}: a Precision@$K$ / NDCG benchmark and an open-vocabulary
query interface would replace the current qualitative,
$20$-tag-bounded evaluation.